\documentclass[11pt, a4paper] {article}
\usepackage[hyperref]{acl2018}
\usepackage[utf8]{inputenc}
\usepackage[misc]{ifsym}
\usepackage{times}
\usepackage{latexsym}
\usepackage{bm}
\usepackage{graphicx}
\usepackage{amsmath}
\usepackage{url}
\usepackage{xcolor} 
\aclfinalcopy
\title{Morphological Word Segmentation on Agglutinative Languages for Neural Machine Translation}
\author{Yirong Pan, \Letter	Xiao Li, Yating Yang and Rui Dong\\
Xinjiang Technical Institute of Physics \& Chemistry, Chinese Academy of Sciences\\
University of Chinese Academy of Sciences\\
Xinjiang Laboratory of Minority Speech and Language Information Processing\\
{\tt\href{mailto:xiaoli@ms.xjb.ac.cn}{xiaoli@ms.xjb.ac.cn}}\\}
\begin{document}
\maketitle
\begin{abstract}
Neural machine translation (NMT) has achieved impressive performance on machine translation task in recent years. However, in consideration of efficiency, a limited-size vocabulary that only contains the top-N highest frequency words are employed for model training, which leads to many rare and unknown words. It is rather difficult when translating from the low-resource and morphologically-rich agglutinative languages, which have complex morphology and large vocabulary. In this paper, we propose a morphological word segmentation method on the source-side for NMT that incorporates morphology knowledge to preserve the linguistic and semantic information in the word structure while reducing the vocabulary size at training time. It can be utilized as a preprocessing tool to segment the words in agglutinative languages for other natural language processing (NLP) tasks. Experimental results show that our morphologically motivated word segmentation method is better suitable for the NMT model, which achieves significant improvements on Turkish-English and Uyghur-Chinese machine translation tasks on account of reducing data sparseness and language complexity.
\end{abstract}
\begin{table*}[t]
\begin{center}
\scalebox{0.93} {
\begin{tabular}{cl}
\hline
Segmentation Strategy & \multicolumn{1}{c}{Sentence Example} \\ 
\hline
Raw & küçük fagernes kasabasındayım , oslo'dan yaklaşık üç saat uzakta . \\
\hline
SCS & küçük fagernes kasaba\#\# sındayım\$\$ , oslo\#\# dan\$\$ yaklaşık üç saat uzak\#\# ta\$\$ . \\
SSS & küçük fagernes kasaba\#\# sı\$\$ nda\$\$ yım\$\$ , oslo\#\# dan\$\$ yaklaşık üç saat uzak\#\# ta\$\$ . \\
BPE & küçük fa@@ ger@@ nes kasaba@@ sın@@ dayım , oslo@@ 'dan yaklaşık üç saat uzakta . \\
\hline
BPE-SCS & küçük fa@@ ger@@ nes kasaba\#\# sındayım\$\$ , oslo\#\# dan\$\$ yaklaşık üç saat uzak\#\# ta\$\$ . \\
BPE-SSS & küçük fa@@ ger@@ nes kasaba\#\# sı\$\$ nda\$\$ yım\$\$ , oslo\#\# dan\$\$ yaklaşık üç saat uzak\#\# ta\$\$ . \\
\hline
English & I am in the small town of F@@ ag@@ er@@ nes , about three hours from Oslo . \\
\hline
\end{tabular}}
\caption{The sentence examples with different segmentation strategies for Turkish-English.}
\label{tab:segmentation example}
\end{center}
\end{table*}
\section{Introduction}
Neural machine translation (NMT) has achieved impressive performance on machine translation task in recent years for many language pairs \citep{Sutskever2014,Bahdanau2015,Luong2015}. However, in consideration of time cost and space capacity, the NMT model generally employs a limited-size vocabulary that only contains the top-N highest frequency words (commonly in the range of 30K to 80K) \citep{Jean2015}, which leads to the Out-of-Vocabulary (OOV) problem following with inaccurate and terrible translation results. Research indicated that sentences with too many unknown words tend to be translated much more poorly than sentences with mainly frequent words. For the low-resource and source-side morphologically-rich machine translation tasks, such as Turkish-English and Uyghur-Chinese, all the above issues are more serious due to the fact that the NMT model cannot effectively identify the complex morpheme structure or capture the linguistic and semantic information with too many rare and unknown words in the training corpus.

Both the Turkish and Uyghur are agglutinative and highly-inflected languages in which the word is formed by suffixes attaching to a stem \citep{Ablimit2010}. The word consists of smaller morpheme units without any splitter between them and its structure can be denoted as ``stem + suffix1 + suffix2 + ... + suffixN''. A stem is attached in the rear by zero to many suffixes that have many inflected and morphological variants depending on case, number, gender, and so on. The complex morpheme structure and relatively free constituent order can produce very large vocabulary because of the derivational morphology, so when translating from the agglutinative languages, many words are unseen at training time. Moreover, due to the semantic context, the same word generally has different segmentation forms in the training corpus.

For the purpose of incorporating morphology knowledge of agglutinative languages into word segmentation for NMT, we propose a morphological word segmentation method on the source-side of Turkish-English and Uyghur-Chinese machine translation tasks, which segments the complex words into simple and effective morpheme units while reducing the vocabulary size for model training. In this paper, we investigate and compare the following segmentation strategies:
\begin{itemize}
\item Stem with combined suffix
\item Stem with singular suffix
\item Byte Pair Encoding (BPE)
\item BPE on stem with combined suffix
\item BPE on stem with singular suffix
\end{itemize}

The latter two segmentation strategies are our newly proposed methods. Experimental results show that our morphologically motivated word segmentation method can achieve significant improvement of up to 1.2 and 2.5 BLEU points on Turkish-English and Uyghur-Chinese machine translation tasks over the strong baseline of pure BPE method respectively, indicating that it can provide better translation performance for the NMT model.
\section{Approach}
We will elaborate two popular word segmentation methods and our newly proposed segmentation strategies in this section. The two popular segmentation methods are morpheme segmentation \citep{Ablimit2010} and Byte Pair Encoding (BPE) \citep{Sennrich2015}. After word segmentation, we additionally add an specific symbol behind each separated subword unit, which aims to assist the NMT model to identify the morpheme boundaries and capture the semantic information effectively. The sentence examples with different segmentation strategies for Turkish-English machine translation task are shown in Table 1.
\subsection{Morpheme Segmentation}
The words of Turkish and Uyghur are formed by a stem followed with unlimited number of suffixes. Both of the stem and suffix are called morphemes, and they are the smallest functional unit in agglutinative languages. Study indicated that modeling language based on the morpheme units can provide better performance \citep{Ablimit2014}. Morpheme segmentation can segment the complex word into morpheme units of stem and suffix. This representation maintains a full description of the morphological properties of subwords while minimizing the data sparseness caused by inflection and allomorphy phenomenon in highly-inflected languages.
\subsubsection{Stem with Combined Suffix}
In this segmentation strategy, each word is segmented into a stem unit and a combined suffix unit. We add ``\#\#'' behind the stem unit and add ``\$\$'' behind the combined suffix unit. We denote this method as SCS. The segmented word can be denoted as two parts of ``stem\#\#'' and ``suffix1suffix2...suffixN\$\$''. If the original word has no suffix unit, the word is treated as its stem unit. All the following segmentation strategies will follow this rule.
\subsubsection{Stem with Singular Suffix}
In this segmentation strategy, each word is segmented into a stem unit and a sequence of suffix units. We add ``\#\#'' behind the stem unit and add ``\$\$'' behind each singular suffix unit. We denote this method as SSS. The segmented word can be denoted as a sequence of ``stem\#\#'', ``suffix1\$\$'', ``suffix2\$\$'' until ``suffixN\$\$''.
\subsection{Byte Pair Encoding (BPE)}
BPE \citep{Gage1994} is originally a data compression technique and it is adapted by \citep{Sennrich2015} for word segmentation and vocabulary reduction by encoding the rare and unknown words as a sequence of subword units, in which the most frequent character sequences are merged iteratively. Frequent character n-grams are eventually merged into a single symbol. This is based on the intuition that various word classes are translatable via smaller units than words. This method making the NMT model capable of open-vocabulary translation, which can generalize to translate and produce new words on the basis of these subword units. The BPE algorithm can be run on the dictionary extracted from a training text, with each word being weighted by its frequency. In this segmentation strategy, we add ``@@'' behind each no-final subword unit of the segmented word.
\begin{table}[t]
\begin{center}
\begin{tabular}{ccc}
\hline
Data & Turkish & English \\
\hline
Sentences & 359,182 & 359,182 \\
Tokens & 6,728,346 & 8,771,170 \\
Vocabulary & 284,252 & 29,443 \\
Stem & 87,770 & - \\
Combined Suffix & 15,722 & - \\
Singular Suffix & 364 & - \\
\hline
\end{tabular}
\caption{The training corpus statistics of Turkish-English machine translation task.}
\label{tab:tren corpus}
\end{center}
\end{table}
\begin{table}[t]
\begin{center}
\scalebox{0.93} {
\begin{tabular}{ccc}
\hline
Data & Uyghur & Chinese \\
\hline
Sentences & 330,192 & 330,192 \\
Tokens & 6,043,461 & 5,947,903 \\
Vocabulary & 261,918 & 43,016 \\
Stem & 128,786 & - \\
Combined Suffix & 25,115 & - \\
Singular Suffix & 224 & - \\
\hline
\end{tabular}}
\caption{The training corpus statistics of Uyghur-Chinese machine translation task.}
\label{tab:uych corpus}
\end{center}
\end{table}
\subsection{Morphologically Motivated Segmentation}
The problem with morpheme segmentation is that the vocabulary of stem units is still very large, which leads to many rare and unknown words at the training time. The problem with BPE is that it do not consider the morpheme boundaries inside words, which might cause a loss of morphological properties and semantic information. Hence, on the analyses of the above popular word segmentation methods, we propose the morphologically motivated segmentation strategy that combines the morpheme segmentation and BPE for further improving the translation performance of NMT.

Compared with the sentence of word surface forms, the corresponding sentence of stem units only contains the structure information without considering morphological information, which can make better generalization over inflectional variants of the same word and reduce data sparseness \citep{Tamchyna2016}. Therefore, we learn a BPE model on the stem units in the training corpus rather than the words, and then apply it on the stem unit of each word after morpheme segmentation.
\subsubsection{BPE on Stem with Combined Suffix}
In this segmentation strategy, firstly we segment each word into a stem unit and a combined suffix unit as SCS. Secondly, we apply BPE on the stem unit. Thirdly, we add ``\$\$'' behind the combined suffix unit. If the stem unit is not segmented, we add ``\#\#'' behind itself. Otherwise, we add ``@@'' behind each no-final subword of the segmented stem unit. We denote this method as BPE-SCS.
\subsubsection{BPE on Stem with Singular Suffix}
In this segmentation strategy, firstly we segment each word into a stem unit and a sequence of suffix units as SSS. Secondly, we apply BPE on the stem unit. Thirdly, we add ``\$\$'' behind each singular suffix unit. If the stem unit is not segmented, we add ``\#\#'' behind itself. Otherwise, we add ``@@'' behind each no-final subword of the segmented stem unit. We denote this method as BPE-SSS. 
\begin{table*}[t]
\begin{center}
\scalebox{0.93} {
\begin{tabular}{cccc}
\hline
Segmentation Strategy & Tokens & Vocabulary & Average Length \\ 
\hline
Raw&6,728,346&284,252&19 \\
\hline
SCS&9,515,077&103,093&26 \\
SSS&11,344,306&87,953&32 \\
BPE&7,692,327&35,047&21 \\
\hline
BPE-SCS&9,887,485&35,315&28 \\
BPE-SSS&11,553,916&34,062&32 \\
\hline
English&8,771,170&29,443&24 \\
\hline
\end{tabular}}
\caption{The training corpus statistics with different segmentation strategies of Turkish}
\label{tab:tr-data}
\end{center}
\end{table*}
\begin{table*}[t]
\begin{center}
\scalebox{0.93} {
\begin{tabular}{cccc}
\hline
Segmentation Strategy & Tokens & Vocabulary & Average Length \\ 
\hline
Raw&6,043,461&261,918&18 \\
\hline
SCS&8,339,079&153,144&25 \\
SSS&9,618,913&128,884&29 \\
BPE&6,712,259&38,288&20 \\
\hline
BPE-SCS&8,976,494&37,814&27 \\
BPE-SSS&9,846,253&39,884&30 \\
\hline
Chinese&5,947,903&43,016&18 \\
\hline
\end{tabular}}
\caption{The training corpus statistics with different segmentation strategies of Uyghur}
\label{tab:uy-data}
\end{center}
\end{table*}
\begin{table*}[t]
\begin{center}
\scalebox{0.93} {
\begin{tabular}{ccccc}
\hline
& \multicolumn{2}{c}{BLEU} & \multicolumn{2}{c}{ChrF3} \\
Segmentation Strategy & test2016 & test2017 & test2016 & test2017 \\
\hline
Raw & 14.5 & 14.0 & 39.4 & 39.3 \\
\hline
SCS & 17.8 & 17.2 & 43.9 & 43.9 \\
SSS&\textbf{18.1}&\textbf{18.2}&44.3& 44.5 \\
BPE& 18.0& 17.5&\textbf{44.9}&\textbf{44.6} \\
\hline
BPE-SCS& 18.7& 17.9&45.4&45.1 \\
BPE-SSS&\textbf{19.2}&\textbf{18.6}&\textbf{46.0}&\textbf{45.5} \\
\hline
\end{tabular}}
\end{center}
\caption{Experimental results of Turkish-English machine translation task.}
\label{tab:tren-results}
\end{table*}
\begin{table}[t]
\begin{center}
\scalebox{0.93} {
\begin{tabular}{ccccc}
\hline
Segmentation Strategy & BLEU & ChrF3 \\
\hline
Raw & 23.5 & 32.2 \\
\hline
SCS & 27.5 & 35.9 \\
SSS & \textbf{27.6} & 36.0 \\
BPE & 27.4 & \textbf{36.6} \\
\hline
BPE-SCS & 29.2 & 37.8 \\
BPE-SSS & \textbf{29.9} & \textbf{38.4} \\
\hline
\end{tabular}}
\end{center}
\caption{Experimental results of Uyghur-Chinese machine translation task.}
\label{tab:uych-results}
\end{table}
\begin{table*}[t]
\begin{center}
\scalebox{0.93} {
\begin{tabular}{cccccc}
\hline
& & \multicolumn{2}{c}{BLEU} & \multicolumn{2}{c}{ChrF3} \\
Merge Operation & Vocabulary & test2016 & test2017 & test2016 & test2017 \\ \hline
10K & 14,832 & 19.0 & 18.2 & 45.6 & 45.4 \\
15K & 21,598 & 18.8 & 18.0 & 45.4 & 45.0 \\
20K & 28,010 & 18.5 & 18.3 & 45.3 & 45.0 \\
\textbf{25K} & 34,062 & \textbf{19.2} & \textbf{18.6} & \textbf{46.0} & \textbf{45.5} \\
30K & 39,828 & 18.7 & 18.4 & 45.3 & 45.1 \\
35K & 45,275 & 18.7 & 18.2 & 45.2 & 45.0 \\ \hline
\end{tabular}}
\end{center}
\caption{Different numbers of merge operations for BPE-SSS strategy on Turkish-English.}
\label{tab:tr-merge-operations}
\end{table*}
\begin{table}[t]
\begin{center}
\scalebox{0.93} {
\begin{tabular}{ccccc}
\hline
Merge Operation & Vocabulary & BLEU & ChrF3 \\
\hline
20K & 24,650 & 28.9 & 38.0 \\
25K & 29,860 & 29.3 & 38.1 \\
\textbf{30K} & 34,925 & \textbf{30.0} & \textbf{38.3} \\
\textbf{35K} & 39,884 & \textbf{29.9} & \textbf{38.4} \\
40K & 44,726 & 28.3 & 37.2 \\
45K & 49,560 & 28.7 & 37.3 \\
\hline
\end{tabular}}
\end{center}
\caption{Different numbers of merge operations for BPE-SSS strategy on Uyghur-Chinese.}
\label{tab:uy-merge-operations}
\end{table}
\section{Experiments}
\subsection{Experimental Setup}
\paragraph{Turkish-English Data :}
Following \citep{Sennrich2016}, we use the WIT corpus \cite{Cettolo2012} and SETimes corpus \citep{Tyers2010} for model training, and use the newsdev2016 from Workshop on Machine Translation in 2016 (WMT2016) for validation. The test data are newstest2016 and newstest2017.
\paragraph{Uyghur-Chinese Data :}
We use the news data from China Workshop on Machine Translation in 2017 (CWMT2017) for model training, validation and test.
\paragraph{Data Preprocessing :}
We utilize the Zemberek \footnote{\url{https://github.com/ahmetaa/zemberek-nlp}} with a morphological disambiguation tool to segment the Turkish words into morpheme units, and utilize the morphology analysis tool \citep{Tursun2016} to segment the Uyghur words into morpheme units. We employ the python toolkits of jieba \footnote{\url{https://github.com/fxsjy/jieba}} for Chinese word segmentation. We apply BPE \footnote{\url{https://github.com/rsennrich/subword-nmt}} on the target-side words and we set the number of merge operations to 35K for Chinese and 30K for English and we set the maximum sentence length to 150 tokens. The training corpus statistics of Turkish-English and Uyghur-Chinese machine translation tasks are shown in Table 2 and Table 3 respectively.
\paragraph{Number of Merge Operations :}
We set the number of merge operations on the stem units in the consideration of keeping the vocabulary size of BPE, BPE-SCS and BPE-SSS segmentation strategies on the same scale. We will elaborate the number settings for our proposed word segmentation strategies in this section.

In the Turkish-English machine translation task, for the pure BPE strategy, we set the number of merge operations on the words to 35K, set the number of merge operations on the stem units for BPE-SCS strategy to 15K, and set the number of merge operations on the stem units for BPE-SSS strategy to 25K. In the Uyghur-Chinese machine translation task, for the pure BPE strategy, we set the number of merge operations on the words to 38K, set the number of merge operations on the stem units for BPE-SCS strategy to 10K, and set the number of merge operations on the stem units for BPE-SSS strategy to 35K. The detailed training corpus statistics with different segmentation strategies of Turkish and Uyghur are shown in Table 4 and Table 5 respectively. 

According to Table 4 and Table 5, we can find that both the Turkish and Uyghur have a very large vocabulary even in the low-resource training corpus. So we propose the morphological word segmentation strategies of BPE-SCS and BPE-SSS that additionally applying BPE on the stem units after morpheme segmentation, which not only consider the morphological properties but also eliminate the rare and unknown words.
\subsection{NMT Configuration}
We employ the Transformer model \citep{Vaswani2017} with self-attention mechanism architecture implemented in Sockeye toolkit \citep{Hieber2017}. Both the encoder and decoder have 6 layers. We set the number of hidden units to 512, the number of heads for self-attention to 8, the source and target word embedding size to 512, and the number of hidden units in feed-forward layers to 2048. We train the NMT model by using the Adam optimizer \citep{Kingma2014} with a batch size of 128 sentences, and we shuffle all the training data at each epoch. The label smoothing is set to 0.1. We report the result of averaging the parameters of the 4 best checkpoints on the validation perplexity. Decoding is performed by beam search with beam size of 5. To effectively evaluate the machine translation quality, we report case-sensitive BLEU score \footnote{\url{https://github.com/EdinburghNLP/nematus/blob/master/data/multi-bleu.perl}} with standard tokenization and character n-gram ChrF3 score \footnote{\url{https://github.com/rsennrich/subword-nmt/tree/master/subword_nmt/chrF.py}}. 
\section{Results}
In this paper, we investigate and compare morpheme segmentation, BPE and our proposed morphological segmentation strategies on the low resource and morphologically-rich agglutinative languages. Experimental results of Turkish-English and Uyghur-Chinese machine translation tasks are shown in Table 6 and Table 7 respectively.
\section{Discussion}
According to Table 6 and Table 7, we can find that both the BPE-SCS and BPE-SSS strategies outperform morpheme segmentation and the strong baseline of pure BPE method. Especially, the BPE-SSS strategy is better and it achieves significant improvement of up to 1.2 BLEU points on Turkish-English machine translation task and 2.5 BLEU points on Uyghur-Chinese machine translation task. Furthermore, we also find that the translation performance of our proposed segmentation strategy on Turkish-English machine translation task is not obvious than Uyghur-Chinese machine translation task, the probable reasons are: the training corpus of Turkish-English consists of talk and news data while most of the talk data are short informal sentences compared with the news data, which cannot provide more language information for the NMT model. Moreover, the test corpus consists of news data, so due to the data domain is different, the improvement of machine translation quality is limited.

In addition, we estimate how the number of merge operations on the stem units for BPE-SSS strategy effects the machine translation quality. Experimental results are shown in Table 8 and Table 9. We find that the number of 25K for Turkish, 30K and 35K for Uyghur maximizes the translation performance. The probable reason is that these numbers of merge operations are able to generate a more appropriate vocabulary that containing effective morpheme units and moderate subword units, which makes better generalization over the morphologically-rich words.
\section{Related Work}
The NMT system is typically trained with a limited vocabulary, which creates bottleneck on translation accuracy and generalization capability. Many word segmentation methods have been proposed to cope with the above problems, which consider the morphological properties of different languages. 

Bradbury and Socher \citep{Bradbury2014} employed the modified Morfessor to provide morphology knowledge into word segmentation, but they neglected the morphological varieties between subword units, which might result in ambiguous translation results. Sanchez-Cartagena and Toral \citep{Toral2016} proposed a rule-based morphological word segmentation for Finnish, which applies BPE on all the morpheme units uniformly without distinguishing their inner morphological roles. Huck \citep{Huck2017} explored target-side segmentation method for German, which shows that the cascading of suffix splitting and compound splitting with BPE can achieve better translation results. Ataman et al. \citep{Ataman2017} presented a linguistically motivated vocabulary reduction approach for Turkish, which optimizes the segmentation complexity with constraint on the vocabulary based on a category-based hidden markov model (HMM). Our work is closely related to their idea while ours are more simple and realizable. Tawfik et al. \citep{Tawfik2019} confirmed that there is some advantage from using a high accuracy dialectal segmenter jointly with a language independent word segmentation method like BPE. The main difference is that their approach needs sufficient monolingual data additionally to train a segmentation model while ours do not need any external resources, which is very convenient for word segmentation on the low-resource and morphologically-rich agglutinative languages.
\section{Conclusion}
In this paper, we investigate morphological segmentation strategies on the low-resource and morphologically-rich languages of Turkish and Uyghur. Experimental results show that our proposed morphologically motivated word segmentation method is better suitable for NMT. And the BPE-SSS strategy achieves the best machine translation performance, as it can better preserve the syntactic and semantic information of the words with complex morphology as well as reduce the vocabulary size for model training. Moreover, we also estimate how the number of merge operations on the stem units for BPE-SSS strategy effects the translation quality, and we find that an appropriate vocabulary size is more useful for the NMT model.

In future work, we are planning to incorporate more linguistic and morphology knowledge into the training process of NMT to enhance its capacity of capturing syntactic structure and semantic information on the low-resource and morphologically-rich languages.
\section*{Acknowledgments}
This work is supported by the National Natural Science Foundation of China, the Open Project of Key Laboratory of Xinjiang Uygur Autonomous Region, the Youth Innovation Promotion Association of the Chinese Academy of Sciences, and the High-level Talents Introduction Project of Xinjiang Uyghur Autonomous Region.
\bibliography{my-paper}
\bibliographystyle{acl_natbib}
\end{document}